# FML-based Prediction Agent and Its Application to Game of Go

Chang-Shing Lee, Chia-Hsiu Kao, Mei-Hui Wang, Sheng-Chi Yang
Dept. of Computer Science and Information Engineering
Knowledge Application and Web Service Research Center
National University of Tainan
Tainan, Taiwan
leecs@mail.nutn.edu.tw

Yusuke Nojima, Ryosuke Saga
Dept. of Electrical Engineering and Information Science
Osaka Prefecture University
Osaka, Japan
nojima@cs.osakafu-u.ac.jp

Nan Shuo, Naoyuki Kubota
Dept. of System Design
Tokyo Metropolitan University
Tokyo, Japan
kubota@tmu.ac.jp

***Abstract*- In this paper, we present a robotic prediction agent including a darkforest Go engine, a fuzzy markup language (FML) assessment engine, an FML-based decision support engine, and a robot engine for game of Go application. The knowledge base and rule base of FML assessment engine are constructed by referring the information from the darkforest Go engine located in NUTN and OPU, for example, the number of MCTS simulations and winning rate prediction. The proposed robotic prediction agent first retrieves the database of Go competition website, and then the FML assessment engine infers the winning possibility based on the information generated by darkforest Go engine. The FML-based decision support engine computes the winning possibility based on the partial game situation inferred by FML assessment engine. Finally, the robot engine combines with the human-friendly robot partner PALRO, produced by Fujisoft incorporated, to report the game situation to human Go players. Experimental results show that the FML-based prediction agent can work effectively.**

## I. Introduction

The game of Go originated from China. There are 381 positions to play for a 19×19 board game [2]. Normally, the weaker player plays Black and starts the game. Stones are consecutively located by two players, Black and White, on the points where the lines cross [3]. Black is also allowed to start with some handicap stones located on the empty board when the difference in strength between the players is large [3]. Moreover, the Go rules contain parts of the ko rule, life and death, suicide, and the scoring method [3]. In the end, the player who controls most territory wins the game. The level of amateur Go players is ranked as Kyu (1K is the highest one) and Dan (1D – 7D and ID is the lowest one). Professional Go players are ranked from 1P to 9P and the highest level is 9P [19]. The strongest professional Go player in the world is Jie Ke from China in Feb. 2017 [20].

Competing with top human players in the ancient, game of Go has been a long-term goal of artificial intelligence [1]. Go's high branching factor makes traditional search techniques or even on cutting-edge hardware ineffective [1]. Additionally, the evaluation function of Go could change drastically with one stone change [1]. Combining a Deep Convolutional Neural Network (DCNN) with Monte Carlo Tree Search (MCTS), Google DeepMind AlphaGo has sent shockwaves throughout Asia and the world since Challenge Match with Lee Sedol in 2016 [3, 5, 6, 19]. Moreover, the new prototype version of AlphaGo played as Master and Magister on the online servers Tygem and FoxGo defeated more than 50 of the top Go players in the world in Dec. 2016 and in Jan. 2017 [7].

The authors would like to thank the financially support sponsored by the Ministry of Science and Technology of Taiwan under the grant MOST 105-2622-E-024-003-CC2 and MOST 105-2221-E-024-017.

Many different real-world applications are with a high-level of uncertainty. A lot of researches proved the good performance of using fuzzy sets. Being an IEEE Standard in May 2016, fuzzy markup language (FML) provides designers of intelligent decision making systems with a unified and high-level methodology for describing systems' behaviors by means of rules based on human domain knowledge [10, 11]. The main advantage of using FML is easy to understand and extend the implemented programs for other researchers. FML is with understandability, extendibility, and compatibility of implemented programs as well as efficiency of programming [10]. There are considerable research applications based on FML and Genetic FML (GFML) such as game of Go [12, 13] and diet [14, 15]. Additionally, Acampora et al. [16] proposed FML Script to make it with evolving capabilities through a scripting language approach. Akhavan et al. [17] used FML-based specifications to validate and implement fuzzy models.

The objective of this paper is to use FML to construct the knowledge base and rule base of the proposed agent and then predict the winner of the game based on the information from darkforest Go open source and extracted sub-games. In this paper, in addition to information provided by darkforest program, we have an additional system to show the information of the game such as current game situation. According to the first-stage prediction results of darkforest Go engine [8, 9] and the second-stage inferred results of the FML assessment engine [18], we further introduce the third-stage FML-based decision support engine to predict the winner of the game. Next, we choose seven games from 60 games (Master vs. top professional Go players in Dec. 2016 and in Jan. 2017) and three games of amateur Go players to evaluate the performance. Finally, we combine playing Go with the fourth-stage robot engine to report real-time situation to Go players. The experimental results show the proposed approach is feasible.

The remainder of this paper is organized as follows: Section II introduces Dynamic DarkForest Go (DyNaDF) cloud platform for game of Go application. Section III describes the proposed FML-based prediction agent by introducing the system structure and the FML-based decision support engine. The experimental results are shown in Section IV. Finally, conclusions are given in Section V.

## II. DyNaDF Cloud Platform for Game of Go

### *A. Structure of Dynamic Darkforest Cloud Platform for Go*

Fig. 1 shows the structure of the DyNaDF Cloud Platform and its brief descriptions are as follows: (1) The DyNaDF cloud platform for game of Go application is composed of a playing-Go platform located at National University of Tainan (NUTN) / Taiwan and National Center for High-Performance Computing (NCHC) / Taiwan, a darkforest Go engine located at Osaka Prefecture University (OPU) / Japan, and the robot PALRO from Tokyo Metropolitan University (TMU) / Japan; (2) Human Go players surf on the DyNaDF platform located at NUTN / NCHC to play with darkforest Go engine located in OPU; (3) The FML assessment engine infers the current game situation based on the prediction information from darkforest and stores the results into the database; (4) PALRO receives the game situation via the Internet and reports to the human Go players; (5) Human can learn more information about game's comments via Go eBook.

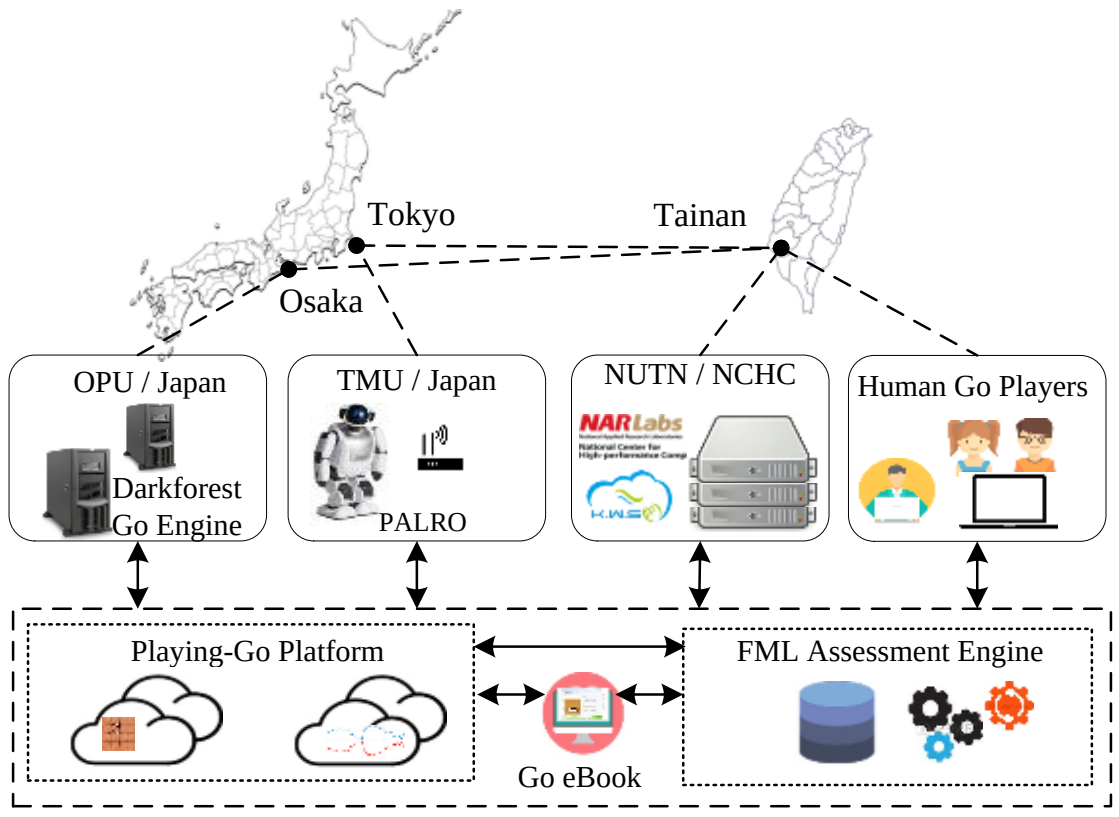

Fig. 1. Structure of dynamic darkforest cloud platform for Go.

## B. *Introduction to Dynamic Darkforest Cloud Platform*

This subsection introduces the developed DyNaDF platform, including a demonstration game platform, a machine recommendation platform, and an FML assessment engine.

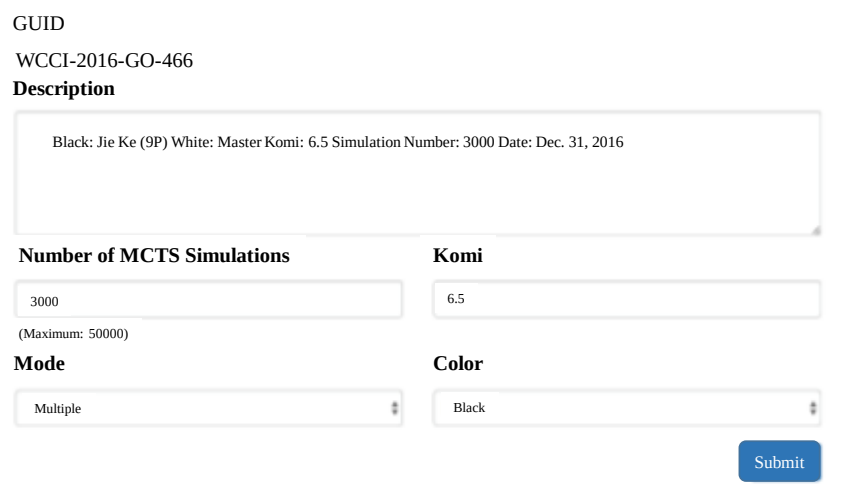

(a)

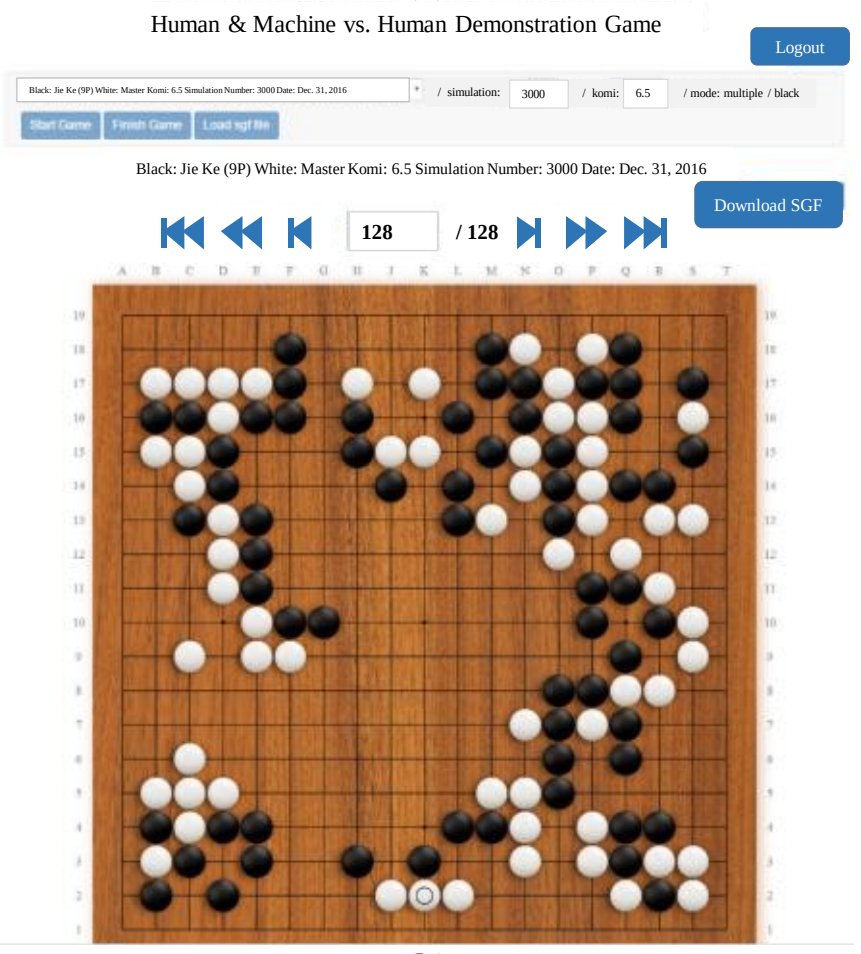

(b)

Fig. 2. Captured screenshots, including (a) setting game and (b) game record provided by the demo game platform.

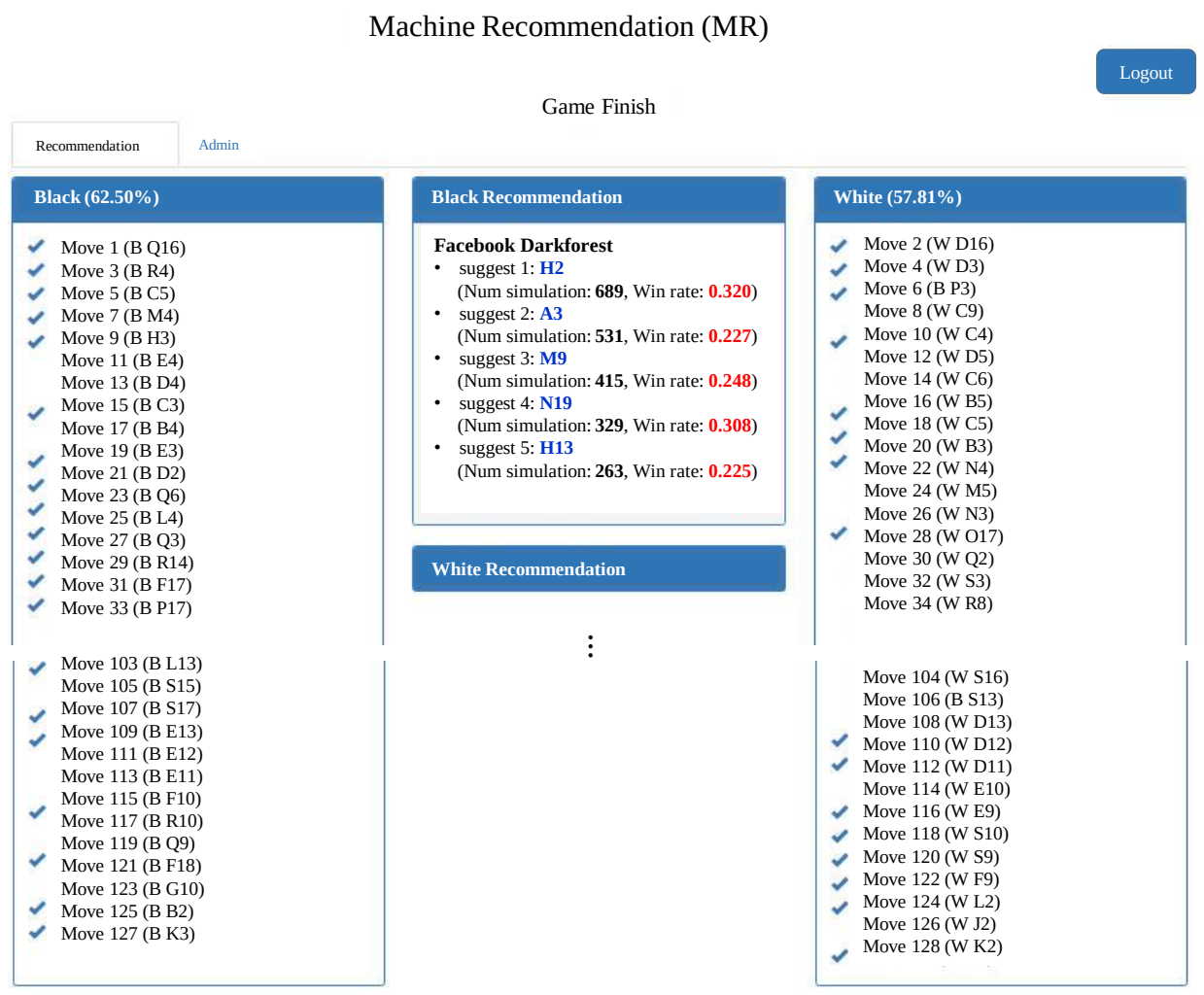

(a)

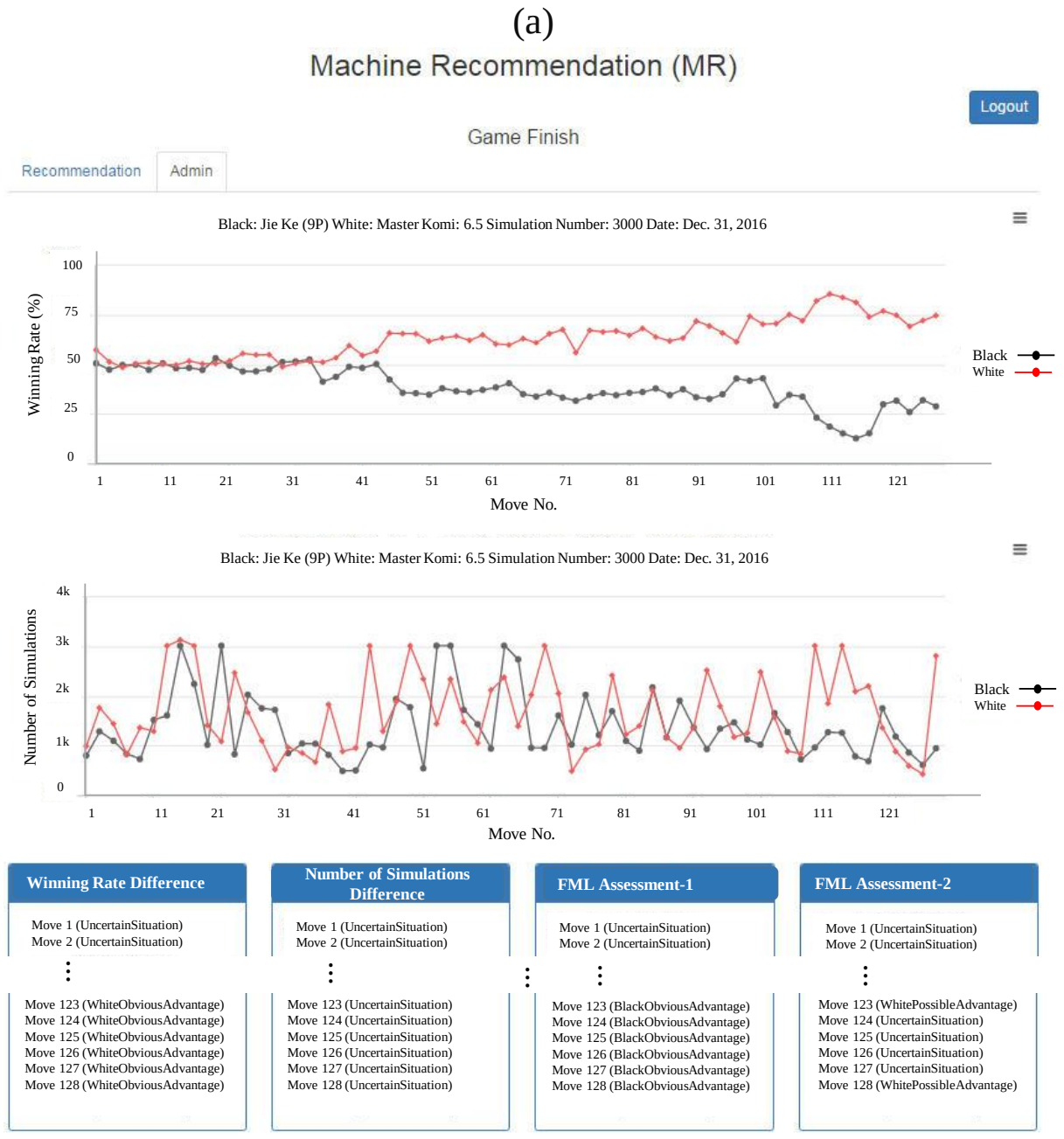

(b)

Fig. 3. Captured screenshots, including (a) predicted next moves and (b) history of winning rrate, the number of simulations, and each-move inferred current game situation provided by the machine recommendation platform.

The game between Jie Ke (9P) as Black and Master as White on Dec. 31, 2016 is used as an example of Figs. 2 and 3. Fig. 2 (a) shows the screen of setting game like the number of MCTS simulations, komi, playing mode, and human is as White or as Black. Fig. 2 (b) shows the game record provided by the demo game platform. Figs. 3 (a)–3 (b) show the predicted next moves as well as history information like winning rate, the number of simulations, and each-move inferred current game situation provided by the machine recommendation platform. FML assessment engine consists of four different methods, including winning rate difference, simulation difference, FML Assessment-1, and FML Assessment-2 [18]. In Fig. 2 (a), we set the number of MCTS simulations to 3000 for darkforest Go engine and acquire the predicted and inferred results from darkforest and the FML assessment engine, respectively. From Fig. 3 (b), we can observe that the top-move rates of Black and White are 62.5% and 57.81%, respectively. Fig. 3 (b) shows that FML Assessment-2 predicts "White is Possible Advantage (WhitePossibleAdvantage)" for the last move.

## III. FML-BASED PREDICTION AGENT

### A. System Structure

Fig. 4 shows the system structure of the proposed four-stage FML-based prediction agent, including *Stage I: darkforest Go engine*, *Stage II: FML assessment engine*, *Stage III: FML-based decision support engine*, and *Stage IV: robot engine*. The followings are its short descriptions:

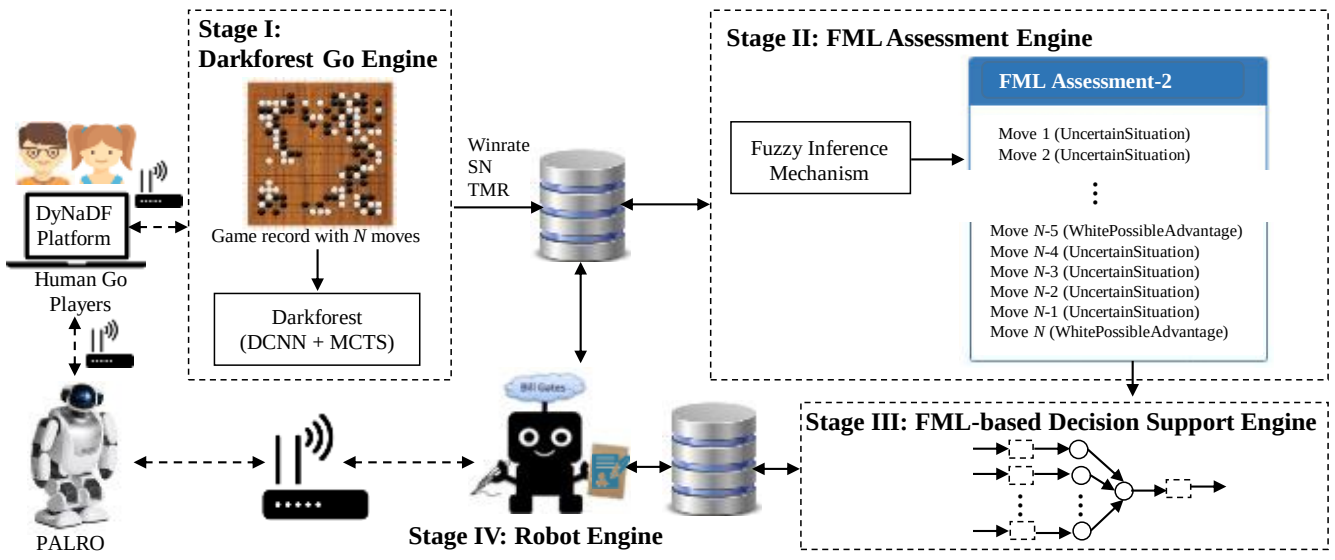

Fig. 4. System structure of four-stage FML-based prediction agent.

- **Stage I:** In this paper, we apply Facebook darkforest Go open source that trains a DCNN to predict the next top-$k$ moves [8]. Darkforest Go engine powered by deep learning has been developed mainly by Tian and Zhu from Facebook AI Research (FAIR) since May 2015 and was open to the public in 2016 [8, 9]. Darkforest has a stable rank of 5D on the KGS servers and pure policy network achieves a stable rank of 3D on KGS. It received the third place and second place in the KGS Go Tournament 2016 and in the ninth UEC Cup Computer 2016, respectively [8, 9]. Darkforest relies on a DCNN designed for long-term predictions and has been able to substantially improve the winning rate for pattern matching approaches against MCTS-based approaches, even with looser search budgets [8]. Tian and Zhu [9] proposed a 12-layered full convolutional network architecture for darkforest where (1) each convolution layer is followed by a ReLU nonlinearity, (2) all layers use the same number of filters at convolutional layers ($w$ = 254) except for the first layer, (3) no weight sharing is used, (4) pooling is not used owing to negatively affecting the performance, and (5) only one softmax layer is used to predict the next move of Back and White to reduce the number of parameters.
- **Stage II:** FML assessment engine adopts each-move-position, darkforest-predicted top-5-move number of simulations and winning rate to decide each-move number of simulations, winning rate, and top-move rate (TMR) [18]. After that, the FML assessment engine infers each-move current game situation, including "*Black is obvious advantage (BlackObviousAdvantage, $B^{++}$)*," "*Black is possible advantage* (*BlackPossibleAdvantage, $B^{+}$*)," "*Both are in an uncertain situation* (*UncertainSituation, U*)," "*White is possible advantage* (*WhitePossibleAdvantage, $W^{+}$*)," and "*White is obvious advantage* (*WhiteObviousAdvantage, $W^{++}$*)."
- **Stage III:** The proposed FML-based decision support engine computes the winning possibility based on the partial game situation inferred by FML assessment engine (see Section III.B) and stores the predicted results into the database.
- **Stage IV:** The robot engine retrieves information from the database to comment on the game situation, including (1) Black and White's move numbers that appear the first 3 highest and the last 3 lowest number of simulations as well as the highest and lowest winning rates, (2) Black and White's average winning rates and top-move rates, and (3) overall game situation. It also reports the real-time predicted top-3-move positions to the human Go player to think carefully before playing his /his next move [18].

### B. FML-based Decision Support Engine

This subsection mainly focuses on the FML-based decision support engine (DSE) whose diagram is shown in Fig. 5.

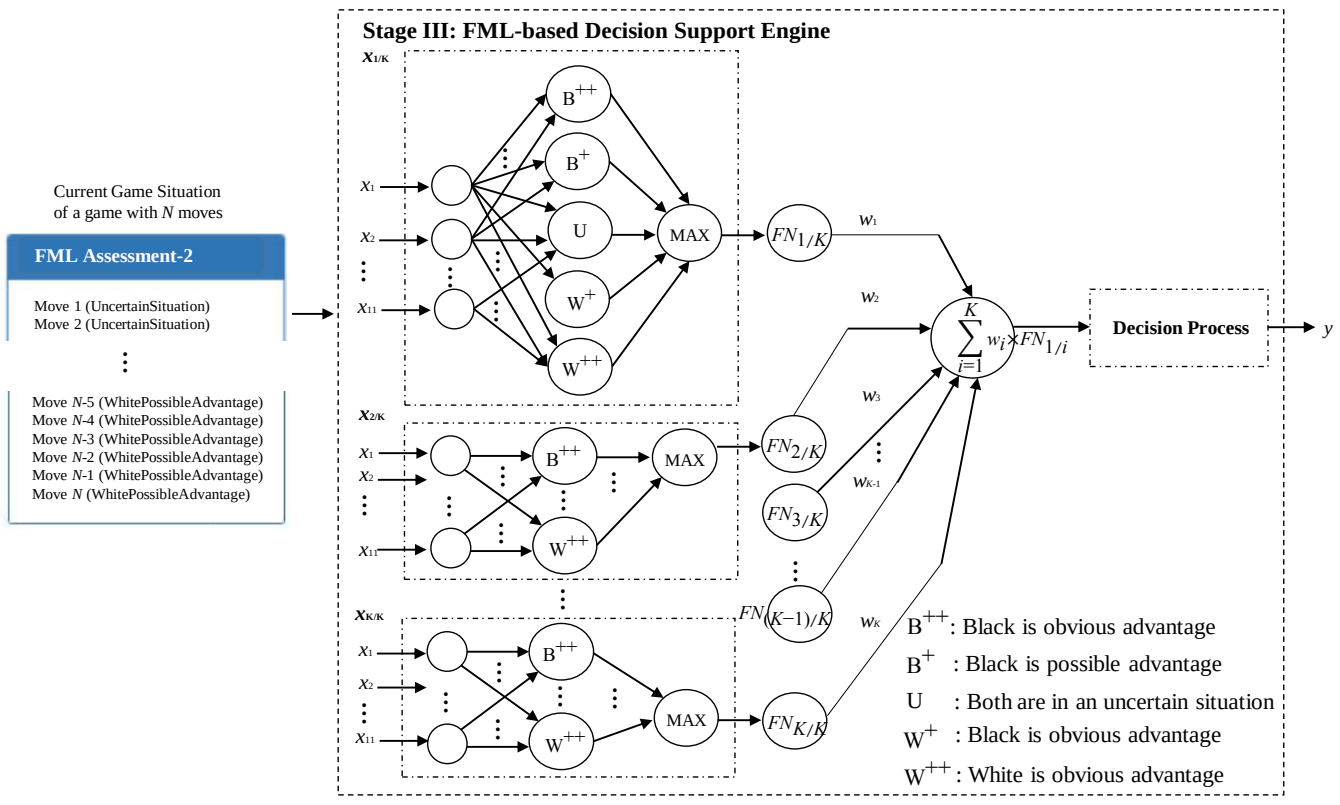

Fig. 5. Diagram of the FML-based decision support engine.

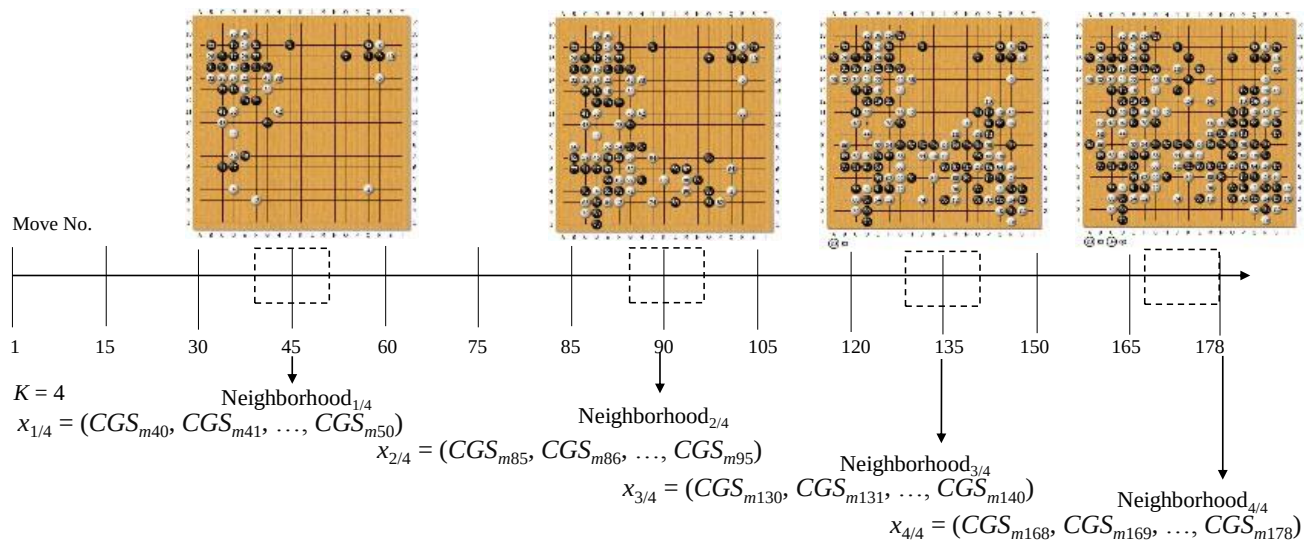

Fig. 6. Example of a game with 178 moves when $K$ = 4.

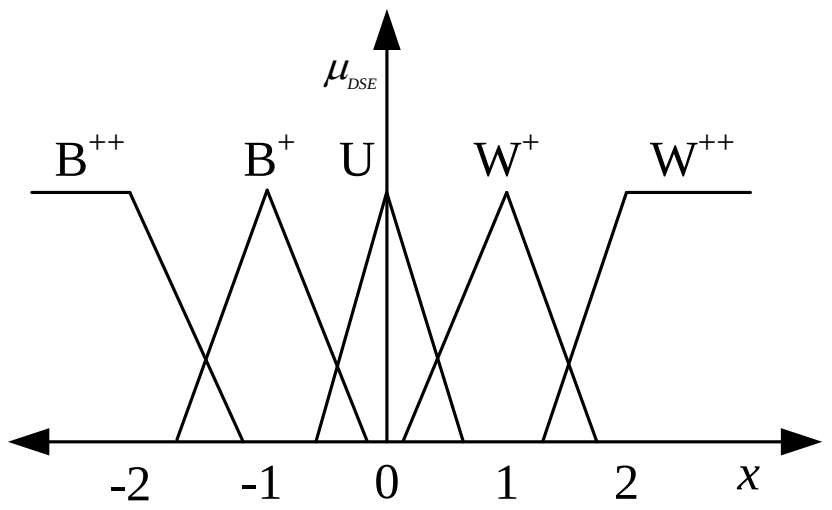

Fig. 7. Fuzzy sets for $B^{++}$, $B^{+}$, U, $W^{+}$, and $W^{++}$.

For example, if the input vector is $\mathbf{x} = (x_1, x_2, \ldots, x_{11})$, then vector $x_{i/k}$ denotes the input 11 current game situations (*CGS*s) in the *Neighborhood*$_{i/k}$. Fig. 6 gives an example of a game with 178 moves when $K$ = 4, where the input information of *Neighborhood*$_{1/4}$, *Neighborhood*$_{2/4}$, *Neighborhood*$_{3/4}$, and *Neighborhood*$_{4/4}$ is the current game situations for *Neighborhood* of fuzzy number ***Move-45*** (moves 40-50), *Neighborhood* of fuzzy number ***Move-90*** (moves 85-95), *Neighborhood* of fuzzy number ***Move-135*** (moves 130-140), and *Neighborhood* of fuzzy number ***Move-Last*** (moves 168-178), respectively. Fig. 7 is the fuzzy sets for $B^{++}$, $B^{+}$, U, $W^{+}$, and $W^{++}$. Table I shows the algorithm of the FML-based DSE.

TABLE I. ALGORITHM OF FML-BASED DECISION SUPPORT ENGINE.

**Input:**
All moves' game situations inferred by FML assessment engine of a game
**Output:**
Game result provided by FML-based decision support engine
**Method:**
**Step 1:** Divide a game into $K$ sub-games
**Step 2:** For $i \leftarrow 1$ to $K$
  **Step 2.1:** $index \leftarrow (N / K)$ /*$N$ denotes the total number of moves of the input game*/
  **Step 2.2:** If $i < K$
    **Step 2.2.1:** Find the current game situations (*CGS*s) for moves [$index \times (i+1) - 5$, $index \times (i+1) + 5$] in the window of $Neighborhood_{i/K}$
  **Step 2.3:** If $i$ equals $K$
    **Step 2.3.1:** Find the *CGSs* for moves [$N$-10, $N$] in the window of $Neighborhood_{K/K}$
  **Step 2.4:** Count individually the number of *CGS*s' linguistics, including $B^{++}$, $B^{+}$, $W^{+}$, and $W^{++}$, in the window of $Neighborhood_{i/K}$ and store the counted results into array $CGS_{ary}$ by
  $CGS_{ary} \leftarrow [count_{B++}, count_{B+}, count_{W+}, count_{W++}]$
  **Step 2.5:** $count_{max} \leftarrow \mathrm{MAX}(CGS_{ary})$ /*Find the maximum of $CGS_{ary}$*/
  **Step 2.6:** $sum \leftarrow \mathrm{SUM}\ (CGS_{ary})$ /*Total the $CGS_{ary}$ */
  **Step 2.7:** If $sum$ equals 0
    **Step 2.7.1:** $FN_{i/K} \leftarrow 0$
  **Step 2.8:** If $sum$ does not equal 0
    **Step 2.8.1:** $CGS_{index} \leftarrow$index of $count_{max}$ /*Set the index of $count_{max}$ to $CGS_{index}$ whose value is 0, 1, 2, or 3*/
    **Step 2.8.2:** $Output_{ary} \leftarrow [-2, -1, +1, +2]$
    **Step 2.8.3:** $FN_{i/K} \leftarrow Output_{ary}[CGS_{index}]$
**Step 3:** $y_{CGS} \leftarrow \sum_{i=1}^{K} FN_{i/K} \times w_i$
**Step 4:** Implement decision process
  **Step 4.1:** If $y_{CGS} \leq -2$
    **Step 4.1.1:** $y \leftarrow B^{++}$
  **Step 4.2:** If $0 > y_{CGS} > -2$
    **Step 4.2.1:** $y \leftarrow B^{+}$
  **Step 4.3:** If $y_{CGS}$ equals 0
    **Step 4.3.1:** $y \leftarrow U$
  **Step 4.4:** If $2 > y_{CGS} > 0$
    **Step 4.4.1:** $y \leftarrow W^{+}$
  **Step 4.5:** If $y_{CGS} \geq 2$
    **Step 4.5.1:** $y \leftarrow W^{++}$
**Step 5:** End

## IV. EXPERIMENTAL RESULTS

### A. Experiment Scenario and Game Information

Fig. 8 shows the scenario of the experiments whose brief descriptions are as follows: (1) the invited human Go players surf on the DyNaDF cloud platform to play with darkforest located in NUTN, NCHC, or OPU, (2) the game records on the Internet are downloaded and fed into the DyNaDF cloud platform, and then (3) the predicted each-move information is generated by darkforest during playing. Table II shows the information of three machines that we installed darkforest Go engine such as machine's location and number & model of GPU. Table III shows the basic profile of the collected games, of which 9 games are from professional Go players, and 4 games from amateur ones. Table IV shows the information of 8 experiments, including the adopted test models (TM1–TM4), collected games (G1–G13), machines that executed darkforest Go engine, namely (OPU, 2), (NUTN, 2), and (NCHC, 4), and the setting of the number of MCTS simulations (500 / 1000 / 5000 / 10000 / 20000).

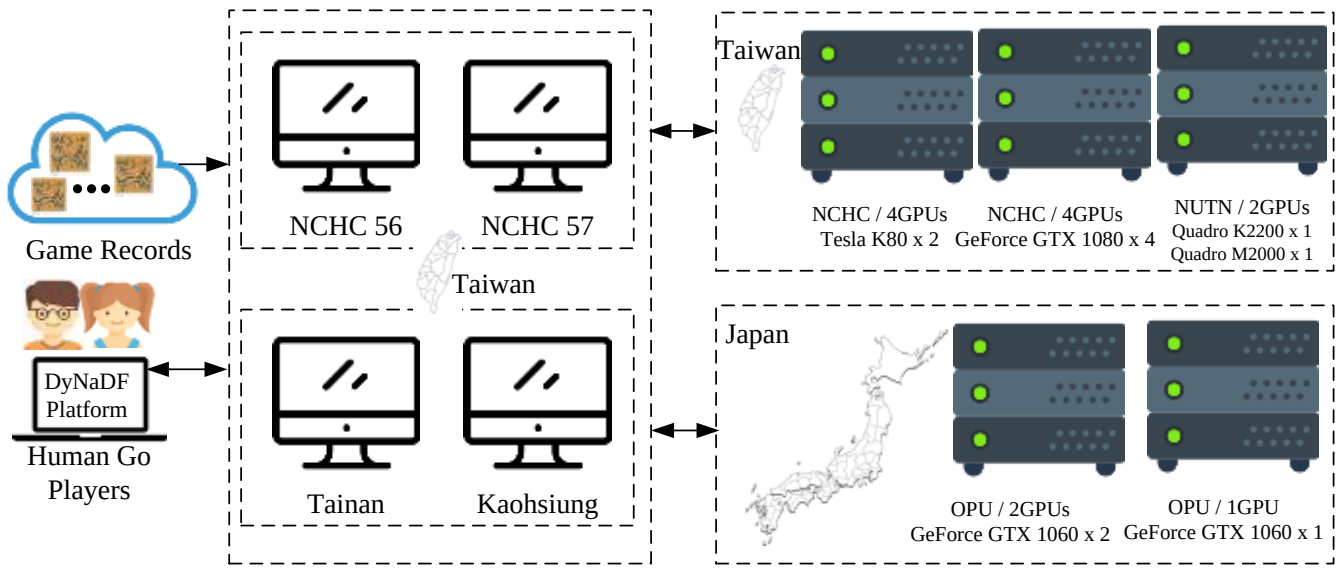

Fig. 8. Scenario of the experiments on DyNaDF platform.

TABLE II. BASIC PROFILE OF ADOPTED MACHINES.

| Location / GPU Number (Abbreviation) | GPU Card Model |
|---|---|
| OPU (Osaka, Japan) / 2 (**OPU, 2**) | GeForce GTX 1060 × 2 |
| NUTN (Tainan, Taiwan) / 2 (**NUTN, 2**) | Quadro K2200×1<br>Quadro M2000×1 |
| NCHC (Tainan, Taiwan) / 4 (**NCHC, 4**) | GeForce GTX 1080 × 4 |

TABLE III. BASIC PROFILE OF COLLECTED 13 GAMES.

| Game No. | Black / Level | White / Level | Date | Winner |
|---|---|---|---|---|
| G1 | Master | Jie Ke / 9P | 2016/12/30 | Black |
| G2 | Jie Ke / 9P | Master | 2016/12/30 | White |
| G3 | Master | Yuta Iyama / 9P | 2017/1/2 | Black |
| G4 | Jie Ke / 9P | Master | 2017/1/3 | White |
| G5 | Chun-Hsun Chou / 9P | Master | 2017/1/4 | White |
| G6 | Master | Weiping Nie / 9P | 2017/1/4 | Black |
| G7 | Li Gu / 9P | Master | 2017/1/4 | White |
| G8 | Shuji Takemura / 1D | Darkforest | 2017/2/9 | White |
| G9 | Minoru Ueda / 5K | Darkforest | 2017/2/9 | White |
| G10 | Lu-An Lin / 6D | Darkforest | 2017/2/20 | Black |
| G11 | Yi-Min Hsieh / 6P<br>Darkforest + PALRO | Chun-Hsun Chou / 9P | 2016/7/25 | White |
| G12 | Yi-Min Hsieh / 6P<br>Darkforest + PALRO | Chun-Hsun Chou / 9P | 2016/7/25 | Uncertain |
| G13 | Lu-An Lin / 6D<br>Darkforest + PALRO | Darkforest | 2016/11/16 | Black |

TABLE IV. INFORMATION OF EXPERIMENTS 1 TO 8.

| Exp. No. | Test Model / Collected Game No. | (Machine Name, No. of GPU) / No. of MCTS Simulations |
|---|---|---|
| Exp. 1 | TM1 / G1–G7 | (**OPU, 2**), (**NUTN, 2**) / 1500 |
| Exp. 2 | TM1 / G1–G7 | (**OPU, 2**), (**NUTN, 2**) / 3000 |
| Exp. 3 | TM1 / G6 | (**OPU, 2**) / 500/5000/10000 |
| Exp. 4 | TM2 / G8 | (**OPU, 2**) / 500 |
| Exp. 5 | TM2 / G9 | (**OPU, 2**) / 1000 |
| Exp. 6 | TM2 / G10 | (**NCHC, 4**) / 20000 |
| Exp. 7 | TM3 / G11–G12 | (**Facebook FAIR**) |
| Exp. 8 | TM4 / G13 | (**NUTN, 2**) / 5000 |

Note:

- Test Model No. 1 (TM1) denotes that we fed the collected game records downloaded from the Internet, for example, Master vs. top professional Go players into our developed DyNaDF cloud platform.
- TM2 denotes that we invited the Go player to play with darkforest via our developed DyNaDF cloud platform.
- TM3 denotes that the weaker human cooperated with PALRO to challenge the stronger human by referring to the recommended top-3-next moves from darkforest.
- TM4 denotes that the human cooperated with PALRO to challenge darkforest by referring to the recommended top-3-next moves from darkforest.

## *B. Darkforest-Predicted Winning Rate and TMR*

The objective of experiments 1 and 2 is to evaluate the variance in winning rate under the different number of MCTS simulations and *Neighborhood* of various fuzzy numbers. The total moves of G1–G7 is 228, 128, 135, 178, 118, 254, and 235, respectively. Fig. 9 shows the darkforest-predicted winning rate for G1–G7 on *Neighborhood* of fuzzy number ***Move-100***, *Neighborhood* of fuzzy number ***Move-200***, and *Neighborhood* of fuzzy number ***Move-Last***. In Fig. 9, the values with a cross (+) denote the human's winning rate of Exp. 2 (OPU, 2) / 3000, while a star (*) is the ones of computer's. Fig. 9 also shows that darkforest successfully predicted that "computer won the game for G1–G7," and "the winning rate difference between computer Go and human is the smallest when G6 was already played 100 moves."

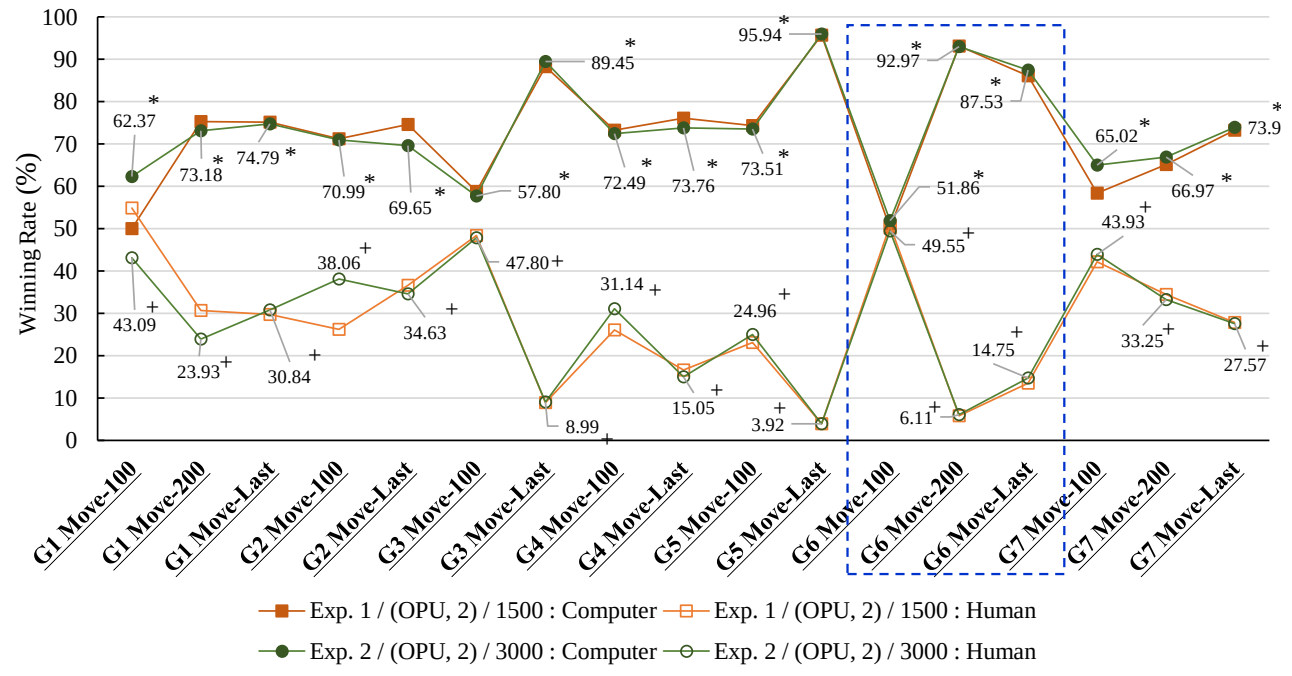

Fig. 9. Winning rate curves for Exps. 1 and 2.

Fig. 10 shows the bar chart of darkforest-predicted winning rate and TMR by two machines, namely (OPU, 2) and (NUTN, 2), when the number of MCTS simulations equals to 1500 and 3000. It indicates that TMR is roughly 50% and there is no obvious difference in TMR when changing the number of MCTS simulations from 1500 to 3000 for both machines (OPU, 2) and (NUTN, 2). Additionally, the winning rates of human and computer are roughly 20%–30% and 70%–80%, respectively, which exactly meets the actual game result (computer won G1–G7). Figs. 11 (a) –11 (b) show the bar charts of winning rate and top-move rate of Exp. 3, respectively. Fig. 11 (a) indicates that the winning rate is still close to each other even human's is higher than computer's at the time of playing move 100. However, the winning rate of computer increases after more moves were played. Observe Fig. 11 (b) that the top-move rate is kept 55%–62% whatever the setting of the number of MCTS simulations is 500, 5000, or even 10000.

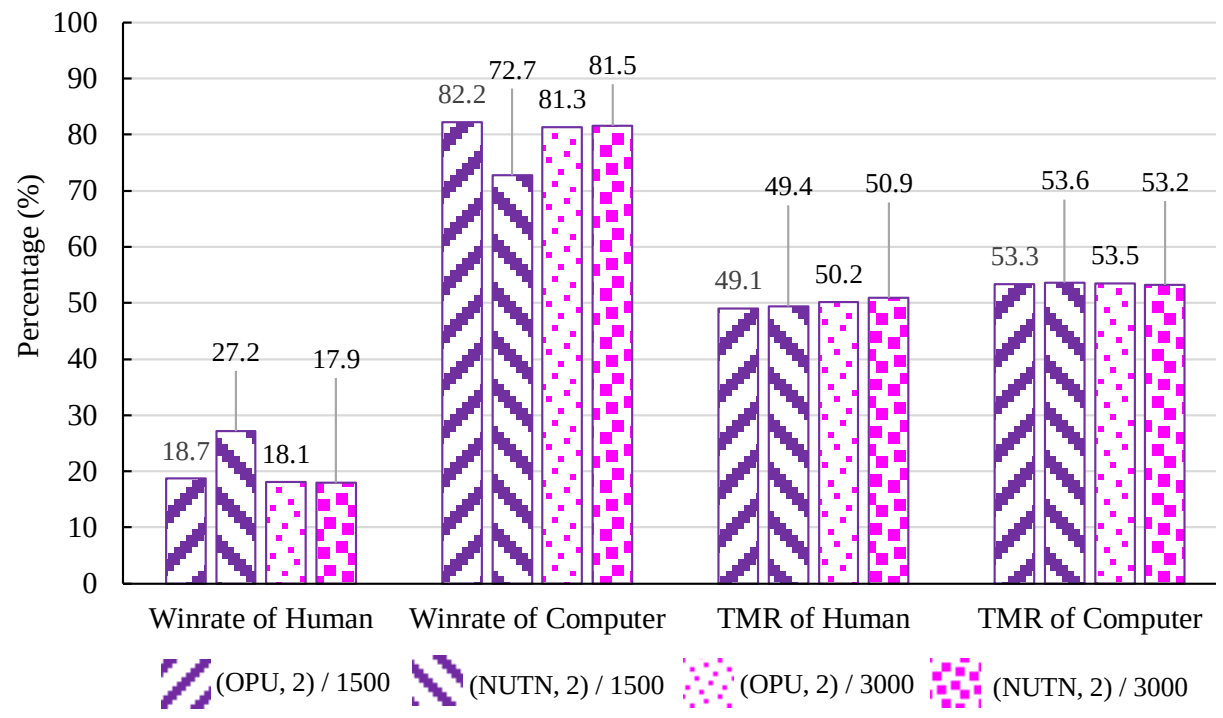

Fig. 10. Bar chart of darkforest-predicted winning rate and TMR by machines (OPU, 2) and (NUTN, 2) for Exps. 1 and 2.

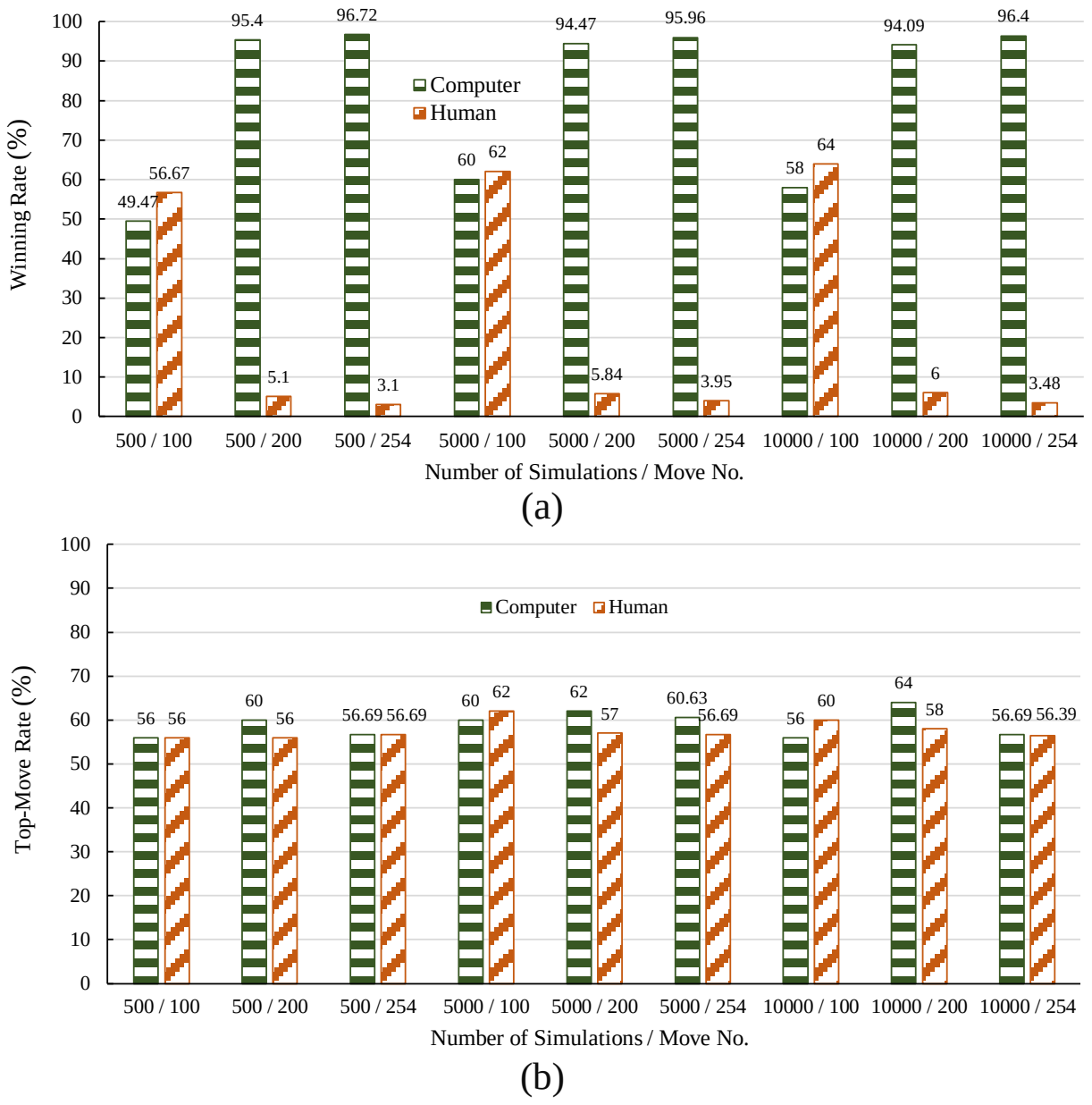

Fig. 11. Bar charts of (a) winning rate and (b) top-move rate for Exp. 3.

## *C. Accuracy of FML-based Decision Support Engine*

This subsection describes the accuracy of FML-based decision support engine. We first used the collected data from Exps. 1 and 2, namely, (OPU, 2) / 1500, (NUTN, 2) / 1500, (OPU, 2) / 3000, and (NUTN, 2) / 3000, to evaluate G1–G7. After that, we inferred the game result based on the current game situations from the last 10 moves to the last one move. The results show each game is correct except G7 of (NUTN, 2) / 1500. Next, we adopted the collected data of G1–G10 from (OPU, 2) / 3000 plus G8–G10 (Exps. 4–6) to evaluate the performance of FML-based DSE. Table V shows the inferred result of each game which indicates that G8 has the problem and the accuracy using $K = 4$ is higher than the one using $K = 3$.

TABLE V. RESULTS OF FML-BASED DECISION SUPPORT ENGINE

| Game No. (Machine Name, No. of GPU) / No. of MCTS Simulations | Winner | $K = 3$ | $K = 4$ |
|---|---|---|---|
| G1: (OPU, 2) / 3000 | Black | $B^+$ | $B^+$ |
| G2: (OPU, 2) / 3000 | White | $W^+$ | $W^+$ |
| G3: (OPU, 2) / 3000 | Black | $B^+$ | $B^+$ |
| G4 (OPU, 2) / 3000 | White | $W^+$ | $W^+$ |
| G5: (OPU, 2) / 3000 | White | $W^+$ | $W^+$ |
| G6: (OPU, 2) / 3000 | Black | $B^+$ | $B^{++}$ |
| G7: (OPU, 2) / 3000 | White | $W^+$ | $W^+$ |
| G8: (OPU, 2) / 500 | White | $\mathbf{B^+}$ | $W^+$ |
| G9: (OPU, 2) / 1000 | White | $W^+$ | $W^{++}$ |
| G10: (NCHC, 4) / 20000 | Black | $B^+$ | $B^+$ |
| Accuracy | | 90% | 100% |

## *D. Human viewpoint on learning with PALRO*

This subsection is to describe the human's viewpoint about Go learning with the PALRO. In Exp. 7, we have two human Go players, 6P Yi-Min Hsieh (Black) and 9P Chun-Hsun Chou (White), who played games with the darkforest and PALRO at the events of IEEE WCCI 2016 and ICIRA 2016. After demonstration game at IEEE WCCI 2016 (for example, G11), Yi-Min Hsieh commented: "*At the critical moment, PALRO will* *cheer for me and sometime it will remind me of having some tea to relax. It seems to have a partner to fight together rather than fight alone. If PALRO can be applied to education or dynamic assessment, it will be helpful for children's learning*." Fig. 12 (a) shows the picture when Yi-Min Hsieh commented on G12 at ICIRA 2016. Fig. 12 (b) shows the picture that Lu-An Lin played G13 with darkforest in Exp. 8. Meanwhile, she also cooperated with darkforest and PALRO reporting the top-3 next moves via Facebook. Lin said that *"Playing with darkforest is helpful and interesting to me. In addition to PALRO's reporting predicted positions, a short time of singing and dancing during the game refreshes me very much."*

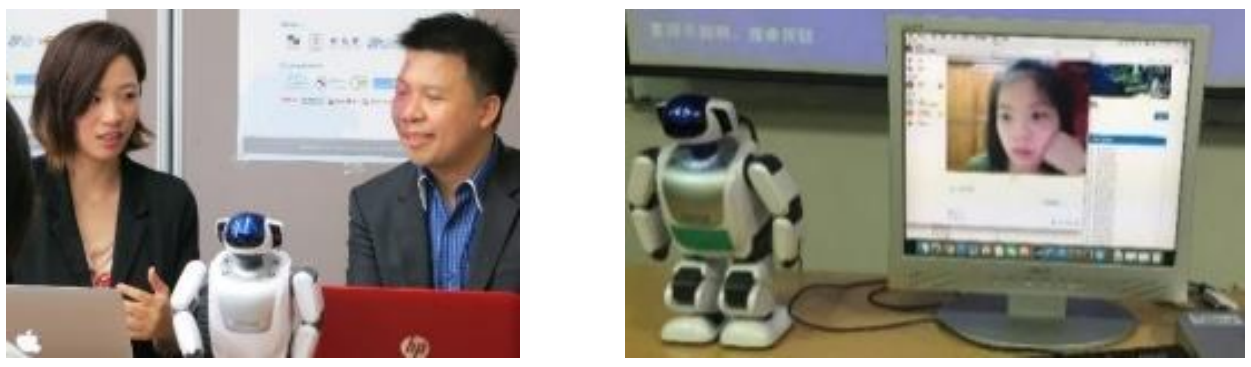

(a) (b)

Fig. 12. (a) Exp. 7: ICIRA 2016 event and (b) Exp. 8: Lu-An Lin played with darkforest and PALRO @ NUTN on Facebook.

## V. Conclusions

This paper presents a robotic prediction agent, including a darkforest Go engine, a fuzzy markup language (FML) assessment engine, an FML-based decision support engine, and a robot engine for game of Go application. The proposed FML-based decision support engine computes the winning possibility based on darkforest's prediction and the partial game situation inferred by FML assessment engine. Additionally, professional Go player commented that "PALRO will be helpful for children's learning if it can be applied to the education." In the future, we will apply machine learning to the proposed platform and collect more data to further evaluate its accuracy and extend the proposed approach to achieve the goal of on-line real-time Go prediction platform. Additionally, the proposed approach will be pertinent to generate play-comments on a game to highlight the positional strategic plan followed by a player during a sequence of moves.

## Acknowledgment

The authors would like to thank all invited Go players for their kind help, especially for Chun-Hsun Chou, Ping-Chiang Chou, Yi-Min Hsieh, Lu-An Lin, Shuji Takemura, and Minoru Ueda. The authors also thank Dr. Yuandong Tian and Yan Zhu for FAIR opening darkforest Go engine. Finally, the authors would like to thank National Center of High Performance Computing (NCHC), especially for Sheng-Hsien Chen, and Kaohsiung City Government, Taiwan for computing resource and technical support.